\documentclass[conference,letterpaper]{IEEEtran} 
\IEEEoverridecommandlockouts

\usepackage[english]{babel}
\usepackage[utf8x]{inputenc}
\usepackage{flushend}
\usepackage{multirow}

\usepackage{xcolor}

\usepackage{amsmath}
\usepackage{amssymb}
\usepackage{graphicx}
\usepackage{gensymb}
\usepackage{subfig}
\usepackage{float}
\usepackage{url,textcomp,listings,xcolor}
\usepackage[colorinlistoftodos]{todonotes}

\usepackage[breaklinks=true,colorlinks,bookmarks=false]{hyperref}
\usepackage{booktabs}
\usepackage{fancyhdr}
\usepackage{xcolor}

\usepackage{tikz}
\usetikzlibrary{calc}

\usepackage{microtype}
\input{glyphtounicode}
\graphicspath{{./figures/}}
\DeclareGraphicsExtensions{.pdf,.jpg,.png,.pdf}

\newcommand{\comment}[1]{}

\newif\ifcolordiff
\colordifftrue

\ifcolordiff
	\newcommand\del[1]{{\color{red}#1}} 
	\newcommand\mvf[1]{{\color{orange}#1}} 
\else
	\newcommand\del[1]{{\iffalse#1\fi}}
	
	\newcommand\mvf[1]{{\iffalse#1\fi}}
	
\fi

\begin{document}
\title{Recovery of Superquadrics from Range Images using Deep Learning: A Preliminary Study}
\author{\IEEEauthorblockN{
Tim Oblak\IEEEauthorrefmark{1}, Klemen Grm\IEEEauthorrefmark{2}, Aleš Jaklič\IEEEauthorrefmark{1},  Peter Peer\IEEEauthorrefmark{1}, Vitomir Štruc\IEEEauthorrefmark{2}, and Franc Solina\IEEEauthorrefmark{1}}
\IEEEauthorblockA{\IEEEauthorrefmark{1}Faculty of Computer and Information Science, University of Ljubljana, Slovenia\\
E-mail: tim.oblak1@gmail.com, \{ales.jaklic, peter.peer, franc.solina\}@fri.uni-lj.si}
\IEEEauthorblockA{\IEEEauthorrefmark{2}Faculty of Electrical Engineering, University of Ljubljana, Slovenia\\
E-mail: \{klemen.grm, vitomir.struc\}@fe.uni-lj.si}
}

\fancypagestyle{plain}{%
  \renewcommand{\headrulewidth}{0pt}
  \chead{This paper was published in IWOBI 2019. }
}

\maketitle
\thispagestyle{plain}
\pagestyle{plain}

\begin{abstract}
It has been a longstanding goal in computer vision to describe the 3D physical space in terms of parameterized volumetric models that would allow autonomous machines to understand and interact with their surroundings. Such models are typically motivated by human visual perception and aim to represents all elements of the physical word ranging from individual objects to complex scenes using a small set of parameters. One of the de facto stadards to approach this problem are superquadrics - volumetric models that define various 3D shape primitives and can be fitted to actual 3D data (either in the form of point clouds or range images). However, existing solutions to superquadric recovery involve costly iterative fitting procedures, which limit the applicability of such techniques in practice. To alleviate this problem, we explore in this paper the possibility to recover superquadrics from range images without time consuming iterative parameter estimation techniques by using contemporary deep-learning models, more specifically, convolutional neural networks (CNNs). We pose the superquadric recovery problem as a regression task and develop a CNN regressor that is able to estimate the parameters of a superquadric model from a given range image. We train the regressor on a large set of synthetic range images, each containing a single (unrotated) superquadric shape and evaluate the learned model in comparaitve experiments with the current state-of-the-art. Additionally, we also present a qualitative analysis involving a dataset of real-world objects. The results of our experiments show that the proposed regressor not only outperforms the existing state-of-the-art, but also ensures a $270\times$ faster  execution time.
\end{abstract}
\section{Introduction}
Artificial intelligence (AI) is inspired by human cognitive abilities and computer vision tries to replicate, at least partially, the functionality of human visual perception. Robots and other artificial systems, on a practical level, need to be aware of the actual 3D physical structure of their surroundings to be able to move around without bumping into obstacles, to grasp, touch and recognize objects, and interact with the physical world. 

Psychologists who study human visual perception agree that at some point the apparent structure of the physical world should be reflected in the perceptions that are formed in the human minds. Biederman~\cite{biederman1987recognition}, a perceptual psychologist, for example, proposed a theory called recognition-by-components, which states that a modest set of volumetric components called geons, can support most recognition tasks by humans. This idea to construct more complex structures from a small set of basic elements is very powerful and is the governing principle in a variety of scientific fields. 
\begin{figure}[!t]
        \centering
        \includegraphics[width=0.5\textwidth]{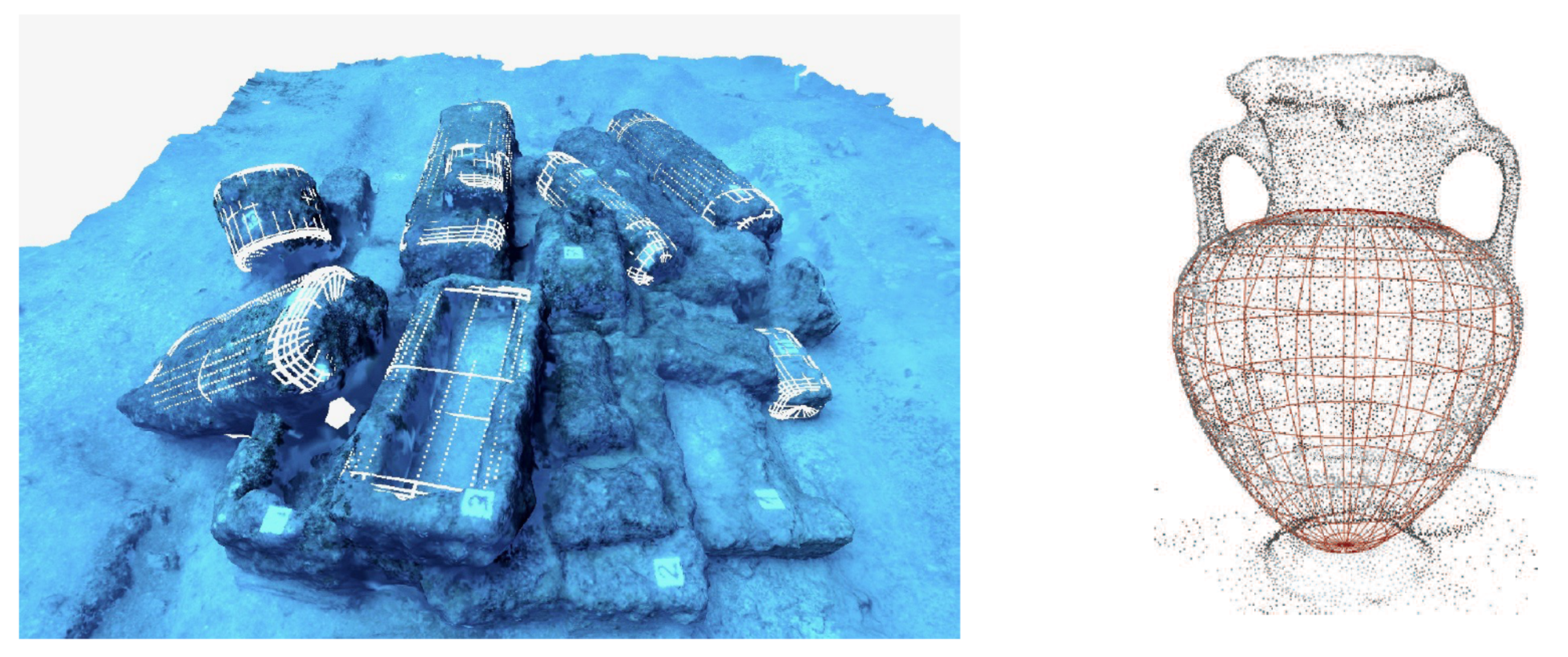}
        \caption{Example use of superquadrics - application in digital heritage. Left: stone blocks and sarcophagi carried by a sunken Roman ship modelled with superquadrics (with courtesy of~\cite{jaklivc2015volumetric}). Right: the body of an amphora, modelled with a superquadric (with courtesy of~\cite{stopinvsek20173d}). In this paper, we try, for the first time, to estimate superquadric representations of 3D shapes from range images using convolutional neural networks (CNNs).  
}
        \label{fig:sq}
\end{figure}

It has been also acknowledged quite early in the progress of computer vision that to mimic human visual perception,
visual information must be at some point represented in terms of spatial or volumetric models since
they can be directly linked to the actual 3D physical space. The search for appropriate  models that could fit this role in computer vision was influenced heavily by Biederman's theory and several options were put forward to act as geons. A particularly popular solution was introduced by Pentland~\cite{pentland86} in the form of superquadrics. Superquadrics are volumetric primitives defined by a closed surface, which can be modeled using a small number of parameters, while still covering a wide variety of 3D shapes, such as ellipsoids, cylinders, parallelopipeds, and all shapes in-between. They can be further extended to represent highly complex 3D structures, as illustrated in Figure~\ref{fig:sq}.    

Several methods for the recovery of superquadrics from range images were already proposed in 80's and 90's~\cite{bajcsy1987three,solina1990recovery} and also included simultaneous segmentation of complex 3D shapes into simpler superquadric-like structures \cite{leonardis1997superquadrics} as well as techniques for geon recognition from the superquadric representations \cite{raja1992recognizing,krivic2004part}. However, a wider application of these methods in computer vision was hindered by their computational complexity caused by the iterative nature of the superquadric-recovery procedures and partially by the difficulty of obtaining high-quality spatial 3D information (suitable for 3D model recovery) at that time.     

With recent advancements in 3D sensing technology, computer vision and most importantly deep learning, it is today possible to devise straight forward solutions for tasks involving optimizations of highly non-linear objective functions that were considered extremely challenging only a few year ago. In line with these trends, we revisit the problem of superquadric recovery in this paper and introduce a deep learning solution for fitting superquadric models to range images. 
Specifically, we design a simple regressor based on a convolutional neural network (CNN) that is able to estimate the parameters of the superquadrics more accuratelly than existing state-of-the-art models and in a fraction of the time. We perform a series of experiments with synthetic, but also real-world images and show that using CNNs for superquadric recovery is a viable option that mitigates many of the shortcomings of earlier techniques. We note, however, that in this preliminary study we only approach a constrained superquadric-recovery problem, where we assume that only a single 3D shape is present in the input data and that the shape can be approximated with an unrotated superquadric model. 

We make the following contributions in this paper:
\begin{itemize}
    \item We present a preliminary study on the use of deep learning models for the recovery of  superquadric models from range images under the assumption that a single 3D shape is present in the input images and that the shape can be represented by an unrotated superquadric.
    \item We introduce a simple CNN-based regressor capable of estimating the parameters of a superquadric model in a fast and efficient manner.
    \item We benchmark the developed CNN regressor against a state-of-the-art method from the literature and report competitive performance in terms of prediction error as well as execution speed.
\end{itemize}

\section{Related Work\label{sec:related}}

In this section we review relevant prior work. We first review existing approaches to the recovery of superquadrics and then discuss closely related literature on CNN models for processing of 3D data. The reader is referred to~\cite{jaklic2000} and~\cite{ioannidou2017deep} for more extensive coverage of these topics. 

\textbf{Superquadric recovery.} Pentland, who introduced supequadrics to computer vision, proposed  to recover them from shading information derived from 2D intensity images \cite{pentland86}.
But this approached proved to be overly complicated and not successful in practice.
Solina and Bajcsy used instead explicit 3D information in the form of range images  \cite{bajcsy1987three,solina1990recovery} 
which are a uniform 2D grid of 3D points as seen from a particular viewpoint.
Solina and Bajcsy 
designed a fitting function that needed to be minimized
to fit a superquadric model to the 3D points.
Since this fitting function is highly non-linear, an iterative procedure could not be avoided for its minimization.

Other researchers have  tried to improve this method in various ways, for example in modifying the fitting function \cite{boult1987recovery},
or using multiresolution~\cite{duncan2013multi} but still essentially relying on iterative methods of minimizing the fitting function.
Instead of gradient least-squares minimization, other methods of minimization have also been tried, such as genetic algorithms \cite{voisin2009genetic}.
Several extensions of superquadrics were proposed in the literature \cite{terzopoulos1990dynamic,hanson1988hyperquadrics},
however, the basic superquadric shape model and the recovery method of Solina and Bajcsy prevailed in most applications of superquadrics, 
in particular for path and grasp planning in robotics, for modelling and interpretation of medical images etc.
Later, Leonardis and Solina expanded Solina and Bajcsy's method to simultaneously 
deconstruct the input range image into several superquadrics, resulting in a perceptually relevant segmentation \cite{leonardis1997superquadrics}. Nevertheless, the procedure still relied on an iterative fitting procedure.

Different from the techniques outlined above, we explore in this study whether recovery techniques relying on contemporary machine learning models, i.e., CNNs, can be used to estimate the parameters of superquadric models without costly iterative optimizations. As we show later, the CNN-based solution described the next section is competitive when compared to state-of-the-art iterative recovery techniques in terms of parameter-prediciton error, but has a significant edge when it comes to processing speed.

\textbf{CNN-based models for 3D visual data.} CNNs have already been employed to process 3D visual data. A CNN regression approach was, for example, used in \cite{7393571} for real-time 2D/3D registration which was, similarly to superquadric recovery, traditionally solved by iterative computation.
CNNs  have also been used to estimate face normals from 2D intensity images instead of standard shape-from-shading methods \cite{trigeorgis2017face} or for fitting 3D morphable models to faces captured in unconstrained conditions \cite{7727386,guler2017densereg,dou2017end,jackson2017large}.

There has been work already on recovering volumetric models using deep neural networks \cite{DBLP:journals/corr/SharmaGF16,wu20153d,Grant2016DeepDR,slabanja2018sq}. Wu et al. \cite{wu20153d}, for example, were building voxel representations of objects, called 3D shapenets, from range images and use CNNs to complete the shape, determine the next best view, and to recognize objects. Sharma et al. \cite{DBLP:journals/corr/SharmaGF16} extend shapenets into full convolutional volumetric auto encoders by estimating voxel occupancy grids. Grant et al. \cite{Grant2016DeepDR} use CNNs to predict volumes on previously unseen image data. Slabanja et al. \cite{slabanja2018sq} deal with superquadric recovery and segmentation of 3D point clouds. 



There is ample evidence by current research that the marriage of 3D data and models relying on the CNN computational paradigm is promising, but only starting. In this preliminary study we add to this body of work by developing a CNN-based solution, which for a selected scene with a single 3D object returns a volumetric description in the form of parameters defining a superquadric model. %

\section{Methodology\label{sec:main}}

In this section we present our approach to the recovery of individual superquadrics with convolutional neural networks (CNNs). We start the section with the introduction of the superquadric-recovery problem and then elaborate on the CNN-model proposed in this work to estimate the parameters of superquadrics from range images. 

\subsection{Problem formulation}

\begin{figure}[!t]
        \includegraphics[width=0.5\textwidth,trim={1mm 23mm 1mm 5mm},clip]{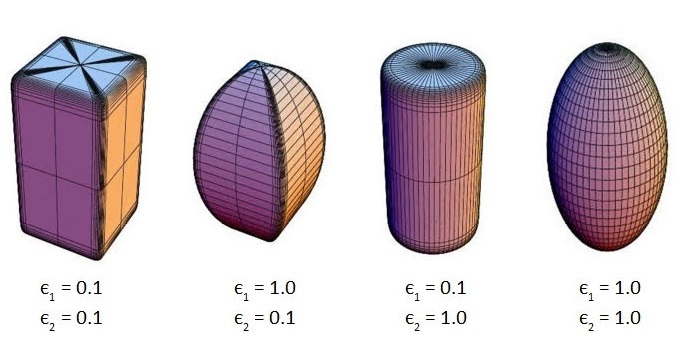}
        \text{\small \hspace{3mm} $\epsilon_1=0.1$ \hspace{11mm} $\epsilon_1=1.0$ \hspace{10mm} $\epsilon_1=0.1$ \hspace{10mm} $\epsilon_1=1.0$}
        \text{\small \hspace{3mm} $\epsilon_2=0.1$ \hspace{11mm} $\epsilon_2=0.1$ \hspace{10mm} $\epsilon_2=1.0$ \hspace{10mm} $\epsilon_2=1.0$}
        \caption{Illustration of a typical family of superquadrics generated with different values of the shape parameters $\epsilon_1$  and $\epsilon_2$.}
        \label{fig:sq-param}
\end{figure}

Superquadrics represents volumetric shapes defined by the following implicit closed surface equation:
\begin{equation}
\label{eq:sq-impl}
F(x, y, z) = \Bigg(\bigg(\frac{x - x_0}{a_1}\bigg)^{\frac{2}{\epsilon_2}}\hspace{-2.8mm}+ \bigg(\frac{y - y_0}{a_2}\bigg)^{\frac{2}{\epsilon_2}}\Bigg)^{\frac{\epsilon_2}{\epsilon_1}}\hspace{-2.8mm}+ \Bigg(\frac{z - z_0}{a_3}\Bigg)^{\frac{2} {\epsilon_1}} 
\end{equation}
where $x_0$, $y_0$ and $z_0$ determine the geometric center of the superquadric $F$ in $\mathbb{R}^3$, $a_1$, $a_2$, and $a_3$ represent the dimensions of the superquadric along each of the coordinate axes, and the parameters ${\epsilon_1}$ and ${\epsilon_2}$ determine the shape of the superquadric. By varying the values of ${\epsilon_1}$ and ${\epsilon_2}$ different 3D shapes can be generated, as illustrated in Figure~\ref{fig:sq-param} \cite{jaklic2000}. The position of the superquadric in a reference coordinate system is defined by its geometric center and the size of the superquadric is determined by the scaling parameters $a_1, a_2$, and $a_3$. For a given point $p=(x, y, z)^{\top}$ in $\mathbb{R}^3$ space, it is possible to determine where it lies in relation to the shape defined by Eq.~\eqref{eq:sq-impl}. If $F(p) = 1$, then the point $p$ lies on the surface of the superquadric, if $F(p) < 1$ then $p$ lies inside of the superquadric, and if $F(p) > 1$, $p$ lies outside the superquadric $F$.  

To account for potential rotations of the superquadrics with respect to the reference coordinate system, a rotation matrix with additional parameters if typically defined over the coordinates of the generated 3D shape. However, in this preliminary study we only consider non-rotated superquadrics and further assume that only a single (isolated) 3D shape is present in the input data. These assumptions make the recovery problem easier as ambiguities due to rotation in 3D are avoided and there is no need for prior segmentation of complex shapes into simpler superquadric-like building blocks.  The recovery problem that we aim to address in this paper can, thus, be defined as a prediction task, where the goal is to estimate the parameters of an (individual/isolated) superquadric model $\mathbf{y}=[a_1, a_2, a_3, \epsilon_1, \epsilon_2, x_0, y_0, z_0]^{\top}\in\mathbb{R}^{1\times8}$ given suitable input data $\mathbf{x}$ (e.g., a range image):
\begin{equation}
    \mathbf{y}=f(\mathbf{x}),
    \label{eq: sq_recovery_problem}
\end{equation}
where $f$ is a predictor that we want to learn from annotated training data. 



\subsection{Model description}

Existing solutions to the recovery of superquadric models typically involve iterative model fitting procedures, which  are computationally expensive and often time-consuming~\cite{jaklic2000}. In this paper, we introduce a novel non-iterative approach that is able to recover the parameters of individual superquadric models from range images without costly iterative optimization techniques. Specifically, we formulate the superquadric-recovery problem as a regression task, where a  convolutional neural network (CNN) is used to predict the parameters of the superquadric model from an input range image $\mathbf{x}$ containing a single 3D shape, i.e.,
\begin{equation}
    \mathbf{y}=f(\mathbf{x};\theta),
    \label{eq: sq_recovery_problem}
\end{equation}
where $\theta$ defines the set of CNN parameters that need to be learned during training.
\begin{figure*}[!t]
        \centering
        \includegraphics[width=0.98\textwidth]{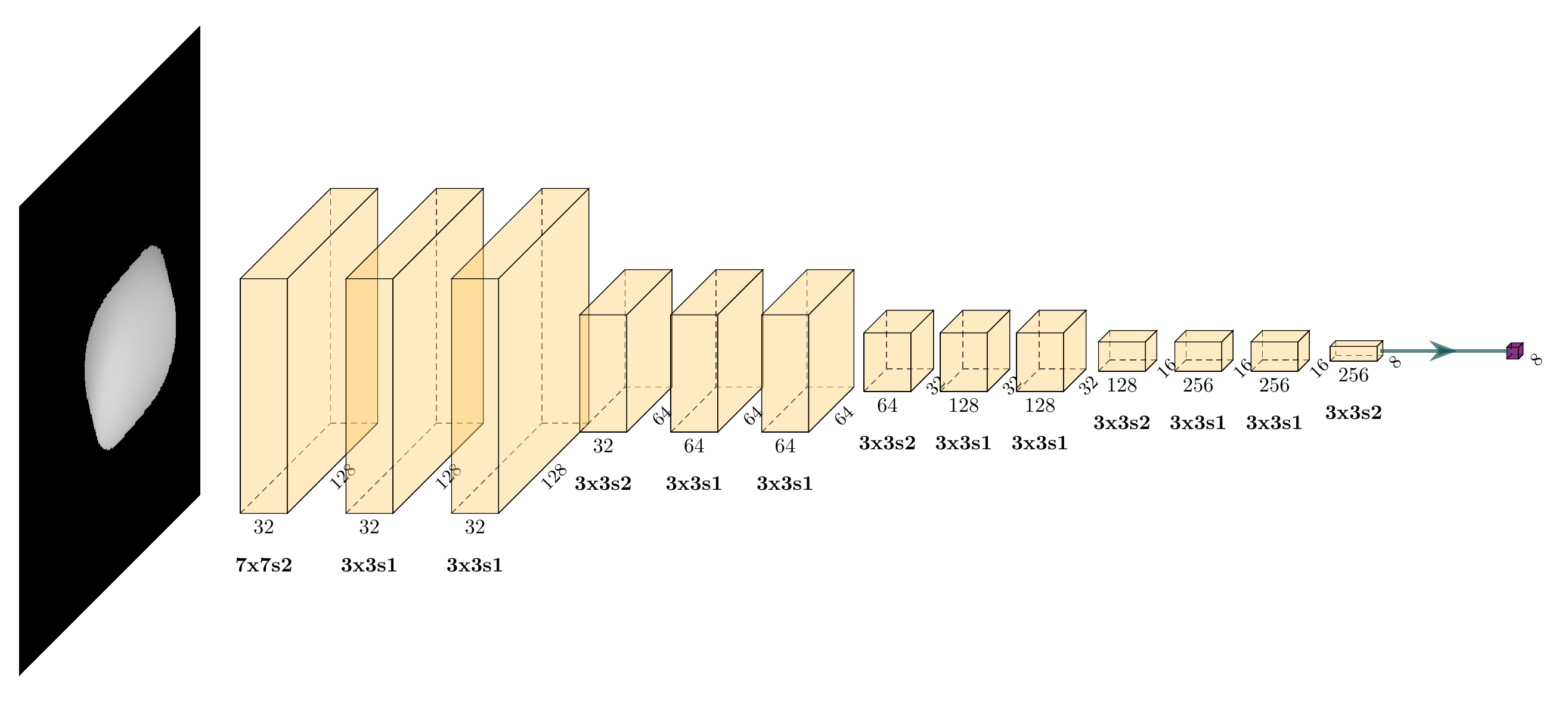}
        \caption{Illustration of the the architecture of the CNN regressor used to predict the parameters of individual superquadrics in this work. We base our model on the VGG network, with decreasing spatial dimensions and increasing filter numbers along the networks layers. Here, we use the $X\times YsZ$ notation to denote a convolutional layer with filters of size $X\times Y$, applied with stride $Z$. The labels on the intermediate representations denote the number of channels and spatial dimensions. Once trained, the model outputs the $8$-dimensional vector of superquadric parameters, defining the position, scale and appearance of the 3D shape in the input range image.}
        \label{fig:cnn-arch}
\end{figure*}

To built our regression model we design a CNN in line with architectural guidelines of the established VGG network~\cite{parkhi2015deep}. We use ideas from the VGG model because it has proven successful in a number of vision-oriented tasks (e.g.,~\cite{emervsivc2018convolutional,emervsivc2017training,grm2017strengths,grm2018face}) and is straight forward to implement. As illustrated in Figure~\ref{fig:cnn-arch}, our model consists of $13$ convolutional layers, each followed by batch normalization and ReLU activation. After the convolutional layers, we add a dense (fully-connected) layer with $8$ real-valued outputs and a linear activation function. The output of the model, thus represents the location, dimension and shape parameters of the superquadric, i.e., the elements of $\mathbf{y}$ from Eq.~\eqref{eq: sq_recovery_problem}. Given the known value range of every parameter, we scale them all to the range $[0, 1]$ before passing the parameters to the model as training targets, so that all outputs follow a similar distribution. To get correct results, we re-scale the model's outputs accordingly at test-time.

Similarly to the VGG model, we gradually decrease the spatial size of the data along the model layers and simultaneously increase the number of channels in the intermediate representations. Due to the particularities of our regression problem, we make a number of additional design choices, i.e.:
\begin{itemize}
    \item The initial convolutional layer in our model has a stride of $2$ and filters of size $7\times 7$ to ensure receptive field coverage. This is because the input range images contain relatively few high-frequency details compared to more general natural images. We, therefore, do not need to process the range data at the full input resolution, and accordingly design the architecture of our model to downsample the intermediate representations more rapidly.
    \item Batch normalization layers are included after every convolutional layer. Based on our preliminary experiments, this significantly reduces overfitting and allows the model to better generalize to unseen images.
    \item Strided convolutions are used instead of the common max-pooling operations for downsampling. As we observed during development, this (slightly) improves the computational efficiency of the model, but does not degrade performance on the regression task.
\end{itemize}


\subsection{Training objective}

We use an $L_2$-norm-based error between the predicted and ground truth superquadric parameters as the training objective for our CNN regressor:
\begin{equation}
\label{eq:mse}
\mathcal{L}\left(\mathbf{y}, \hat{\mathbf{y}}\right) =|| \mathbf{y} - \hat{\mathbf{y}}||_2^2
\end{equation}
where $\mathbf{y}$ represents the ground truth parameters, $\hat{\mathbf{y}}$ represents the predictions of the CNN model, i.e., $\hat{\mathbf{y}} = f(\mathbf{x}; \theta)$, $f$ is the CNN model with parameters $\theta$, and $\mathbf{x}$ is the input range image.

To learn the CNN regressor we use the ADAM minibatch stochastic gradient descent optimization algorithm and minimize the loss over the available training data to find the parameters of the CNN regressor, ${\theta}$. 


\section{Experiments\label{sec:experiments}}
In this section we present experiments aimed at analyzing the performance of the proposed CNN regressor. We start the section with a description of the experimental dataset and model training procedure and then proceed to the results and corresponding discussion. 

\subsection{Dataset, experimental setup and performance metrics}

\textbf{Experimental dataset and setup.} To train and evaluate the CNN regressor presented in the previous section, we generate a synthetic dataset of 3D shapes and render them in the form of range images. Generating synthetic superquadrics allows us to create considerable amounts of data quickly and ensures that each computer generated image has a corresponding ground truth (i.e., superquadric parameters) needed for both, training and testing. 

We generate the data in a controlled manner using a custom rendering tool, where superquadric models with arbitrary parameters can be created. Each range image in the dataset contains a single superquadric inside a $256\times256\times256$ grid, where the first two dimensions encode the image width and height and the last dimension encodes the depth, resulting in a 3D range image. Higher pixel values correspond to closer ranges, while pixels with zero values correspond to the background. The renderer accepts a total of $17$ arguments: positional parameters ($x_0$, $y_0$ and $z_0$ coordinates of the geometric centre), shape parameters ($\epsilon_1$  and $\epsilon_2$), dimension parameters ($a_1$, $a_2$ and $a_3$) and rotational parameters (elements of a $3\times3$ rotation matrix). The rotational parameters are not used in this work, as mentioned earlier. 

The parameters supplied to the renderer are constrained within specific ranges, as the generated superquadrics need to reside inside a $256\times256\times256$ grid. The parameters defining the position and size of the superquadrics can, therefore, only take values from the interval $[0,256]$ and the parameters determining the shape of the superquadrics can only take values from the interval $[0,1]$. However, in practice we also want the entire 3D shape to be visible and therefore use somehow narrow parameter ranges when generating our dataset. Specifically, we define the geometric centre of the 3D shape in each image by drawing coordinates independently from the distribution $\mathcal{U}\left(25, 230\right)$, the dimensions (size) of the superquadric for each of the axes by sampling from the distribution $\mathcal{U}\left(25, 75\right)$ and the shape parameters, $\epsilon_1$ and $\epsilon_2$, drawing values independently from the distribution $\mathcal{U}\left(0.1, 1\right)$. As shown in the Figure \ref{fig:sq-param}, this distribution of parameters results in a diverse set of shapes that include cuboids, ellipsoids and cylindrical shapes. These shapes and the corresponding ground truth parameters can then be used for training and testing of the proposed CNN regressor.

We generate a total of $M=100,000$ range images for training and $N=20,000$ range images for the performance assessment. Note that all generated images are rendered in an isometric projection, so that multiple sides of the superquadrics are always visible in every image.
\begin{figure}[t]
        \centering
        \includegraphics[width=0.5\textwidth,trim={3mm 2mm 1mm 2mm},clip]{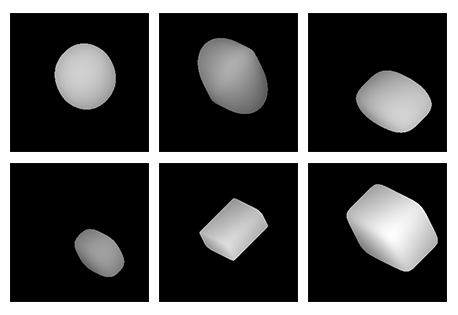}
        \caption{Sample images from the synthetic dataset. Each generated range image contains a single superquadric, which is rendered in an isometric projection. The entire dataset contains a diverse set of superquadrics including cuboids, ellipsoids and cylindrical shapes.}
        \label{fig:data-synth}\vspace{-1mm}
\end{figure}



\textbf{Performance metrics.} To evaluate the performance of our CNN model, we report the Mean Absolute Error (MAE) for each of the estimated superquadric parameters $\rho\in\{a_1, a_2, a_3, \epsilon_1, \epsilon_2, x_0, y_0, z_0 \}$ in our experiments:
\begin{equation}
\label{eq:mae}
MAE = \frac{1}{N}\sum_{i=1}^{N}  | \rho_i - \hat{\rho}_i|
\end{equation}
where $\rho_i$ and $\hat{\rho}_i$ are the ground truth and predicted parameter values for the $i$-th image in the test set, respectively, and $N$ denotes the total number of test images used in the experiments.




\subsection{Model training}

As indicated earlier, we use a training set of $100,000$ synthetically rendered superquadric range images to learn the parameters of the proposed CNN regression model. We split the training dataset into a set of $80,000$ images that are  used for the actual learning procedure and a set of $20,000$ validation images that are used to observe the generalization abilities of the model during training. We initialize the parameters of the CNN regression model using the uniform distribution method proposed by~\cite{He_2015_ICCV}, and train it using the Adam~\cite{kingma2014adam} stochastic gradient optimization algorithm. We use a batch size of $256$, and an initial learning rate of $10^{-3}$, which is successively reduced by a factor of $10$ at the end of epochs $250$ and $500$. We use the MSE loss over the validation dataset as our stopping criterion, with a patience factor of $15$ epochs. Since we have sufficient training data available, we do not perform any data augmentation. The model converges after around $600$ epochs. The training was done on a NVIDIA GTX Titan XP GPU, and took approximately $36$ hours to complete.

Once trained, the model takes a $256\times 256$ range image as input and returns the 8-dimensional vector of superquadric parameters $\mathbf{y}$ at the output.

\subsection{Results and discussion}
We now present results of the experimental evaluation of the proposed CNN model. 

\textbf{Model evaluation.}
In the first series of experiments, we assess the performance of the proposed CNN regressor using the generated $20,000$ test images from our synthetic dataset. For comparison purposes we also include results for the state-of-the-art iterative approach to superquadric recovery from Solina and Bajcsy~\cite{solina1990recovery}. The results of the assessment are presented in Table~\ref{table:results-iso} in the form of MAE scores for each of the $8$ superquadric parameters considered and average processing times required to process a single input image - computed over $20,000$ test images. 
\begin{table*}[!t]
\caption{Performance assessment and comparison with the state-of-the-art. The table shows MAE scores for each of the 8 superquadric parameters and the average processing time needed to estimate the parameters from a single input image. Note that the proposed CNN-based approach achieves comparable error metrics as the state-of-the-art approach from Solina and Bajcsy~\cite{solina1990recovery}, but ensures a speedup of more than $270\times$. The values in the square brackets indicate the possible range for each parameter group.}
    \begin{center}
    \renewcommand{\arraystretch}{1.20}
    \begin{tabular}{lccccccccccccr}
    \hline\hline
    \multirow{ 2}{*}{Approach} & & \multicolumn{3}{c}{Dimensions [0-256]} & & \multicolumn{3}{c}{Position [0-256]} & & \multicolumn{2}{c}{Shape [0-1]}& & \multirow{ 2}{*}{Processing time} \\\cline{3-5} \cline{7-9} \cline{11-12} 
    & & $a_1$ & $a_2$ & $a_3$ & &  $x_0$ & $y_0$ & $z_0$ & &  $\epsilon_1$ & $\epsilon_2$ & \\
    \hline
    CNN regressor (ours) & & $\mathbf{1.014}$ & $\mathbf{1.024}$ & $\mathbf{0.965}$ & & $\mathbf{0.703}$ & $0.859$ & $\mathbf{1.834}$ & & $\mathbf{0.015}$ & $\mathbf{0.018}$ & & $\mathbf{3.6}$~ms\\
    Solina and Bajcsy~\cite{solina1990recovery} & & $7.958$ & $7.461$ & $12.315$ & & $0.744$ & $\mathbf{0.745}$ & $1.888$ & & $0.155$ & $0.279$ & & $988.9$~ms \\
    \hline\hline
    
    \end{tabular}
    \label{table:results-iso}
    \end{center}
\end{table*}
\begin{figure*}[!t]
        \centering
        \includegraphics[width=0.9\textwidth,trim={3mm 4mm 3mm 5mm},clip]{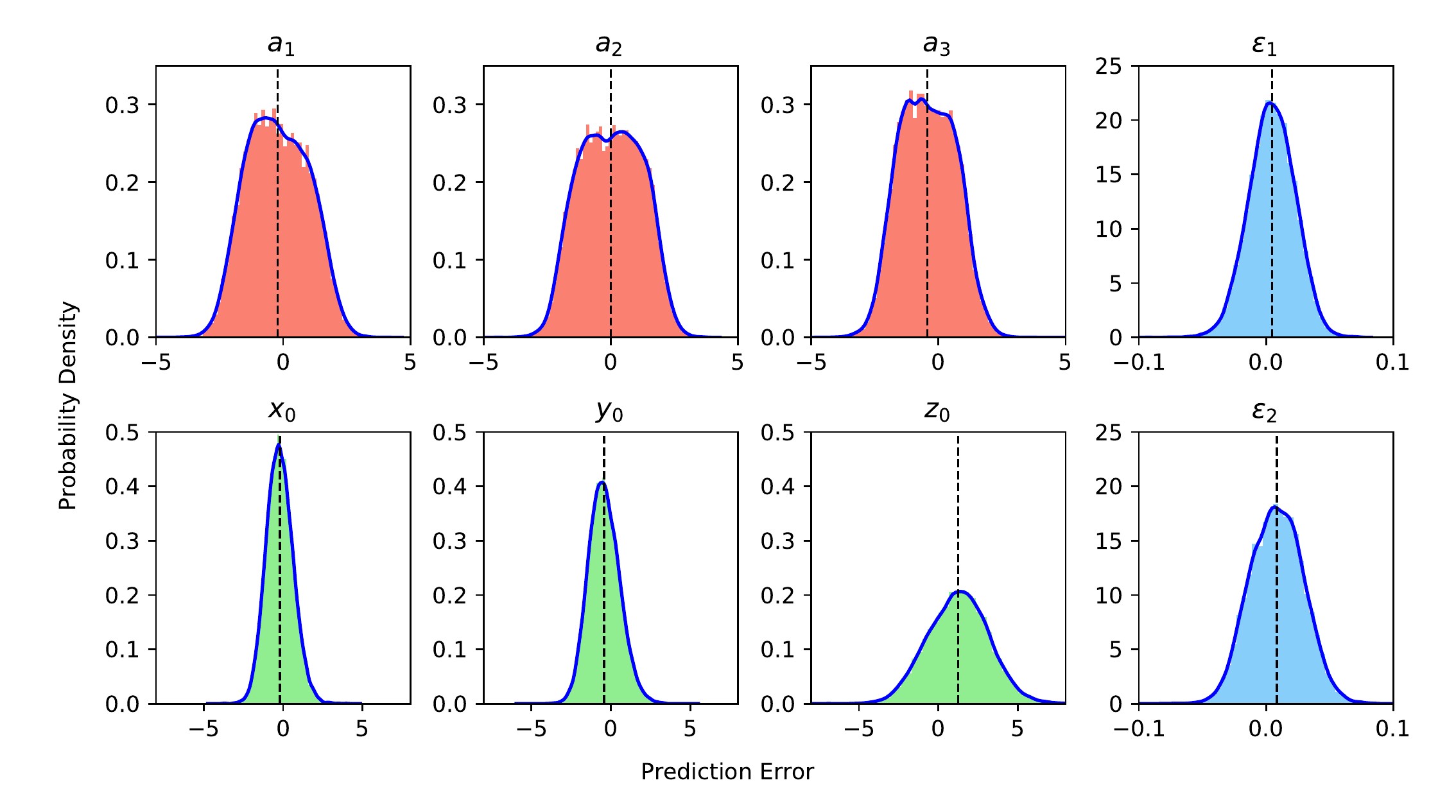}
        \caption{Comparison of the error distributions for the $8$ predicted parameters with the proposed CNN-based model. Note that the distributions are relatively narrow with the majority of the mass located close the mean. This shows that even for more challenging superquadrics the predictions are close to the true values. The figure is best viewed in color.}
        \label{fig:results-hist}
\end{figure*}

As can be seen, the average MAE scores for the proposed CNN-based approach are relatively small for all $8$ parameters compared to the possible range of parameter values. The estimates of the dimension (or scale) parameters $a_1$, $a_2$ and $a_3$, for example, all have a mean absolute error of around $1$, which accounts for approximately $0.39\%$ of the available parameter range. The predicted positional parameters $x_0$, $y_0$ and $z_0$ displays a more inconsistent behavior with MAE scores ranging from the smallest of $0.703$ for $x_0$ to largest of $1.834$ for $z_0$. Interestingly, while the $x$ and $y$ coordinates of the geometric center of the superquadric exhibit a similar MAE value, the estimate of the $z$ coordinate exhibits an error twice as large. Nonetheless, even considering the largest of the errors on the positional parameters, the center of the superquadric is still estimated incorrectly by less than $2$ voxels in the $256\times256\times256$ grid on average.  Similar results are also observed for the shape parameters $\epsilon_1$ and $\epsilon_2$ with average absolute errors below $2\%$ of the available parameter range.         

When comparing the results of the proposed CNN regressor to the state-of-the-art method from Solina and Bajcsy, we see a significant improvement in performance with most of the parameters. Our method improves the prediction of dimensional parameters by $3.6\%$ and the prediction of shape parameters by $20\%$ relative to the corresponding parameter ranges on average. The positional parameters for the iterative methods have comparable results to our CNN regressor and share the same irregular behavior, with prediction errors for the parameter $z_0$ being twice as large as the errors for $x_0$ and $y_0$. In absolute terms, the CNN regressor is able to reduce the prediction error by around one order of magnitude for the dimensional ($a_1$, $a_2$ and $a_3$) and shape ($\epsilon_1$,$\epsilon_2$) parameters and produces comparable positional parameters ($x_0$, $y_0$ and $z_0$). 

Another major contribution are significantly faster processing times. Our approach requires $3.6$ ms on average to process a single input image, whereas the iterative method from Solina and Bajcsy needs $988.9$ ms. Thus, our model is able to achieve state-of-the-art prediction performance, but ensures a speed up of more than $270\times$. For a fair interpretation of the results, it needs to be noted that our CNN regressor is able to make use of GPUs during processing, while the competing method needs to run on a CPU (Intel Core i7-8550U in our case) due to its iterative nature. 



%
    
    
    


\textbf{Model analysis.} The results reported in the previous section show only a partial picture of the performance of the proposed CNN regressor. Since the synthetic images in our dataset have different characteristics it makes sense to evaluate how the prediction errors are distributed across the test images. To this end, we show in Figure~\ref{fig:results-hist} error probability density functions for each of the $8$ predicted parameters. Note that the graphs show the distribution for the prediction errors and not for the absolute errors, so the errors may also be negative.

We observe that all distributions have a close-to-Gaussian shape with most of the mass located close to the mean. This observation suggests that even for difficult images, the parameter estimates produced by our CNN regressor are reasonable approximations of the true values.

Another important observation we can make from the presented densities is related to model bias. Since our CNN regressor is a statistical model, examining the model bias is useful for determining the most likely parameter values the regressor will predict. If we look at the mean error values marked by the vertical dotted line in each graph, we observe that for the shape parameters, $\epsilon_1$ and $\epsilon_2$, the model exhibits a small bias toward positive errors on average, which means that an object, rendered from the predicted parameters will be slightly more rounded compared to the original object. Among the coordinates of the geometric center of the superquadric predicted by our CNN regressor, the $z_0$ coordinate exhibits a somewhat larger bias than the $x_0$ and $y_0$ coordinates. The error for the $z_0$ coordinate is biased toward positive prediction errors on average,  which suggests that an object, generated from the predicted parameters would appear closer in the scene in relation to the original. This observation could be a case against the usage of classic convolutional filters, which appear to be spatially aware in the $x$ and $y$ directions, but lack depth perception on the $z$ coordinate axis. This observation is also supported by the significantly larger standard deviation of the prediction error for the $z_0$ coordinate, which is $1.92$ in comparison to the standard deviations of $0.85$ and $0.97$ for the $x_0$ and $y_0$ coordinate, respectively. The dimensional/scale parameters $a_1$ and $a_2$ show little bias with an average prediction error close to $0$, the scaling factor $a_3$ along the $z$ axis, on the other hand, exhibits a small bias toward negative values, suggesting that the rendered superquadrics would be slightly smaller than the original 3D shapes along the depth dimension on average.      

\textbf{Performance with real data.} To evaluate our CNN regressor on real-world data, we collect a small dataset of 3D shapes using an Artec MHT 3D scanner. The scanner is designed to capture clouds of 3D points and a wide spectrum of colors (up to 24 bpp). Capturing both color information of the object’s appearance and its geometry results in textured high-quality 3D  models, which can then be processed by a number of 3D software packages. 

\begin{figure}[!t]
        \centering
        \includegraphics[width=0.48\textwidth,trim={3.3mm 2mm 1mm 3mm},clip]{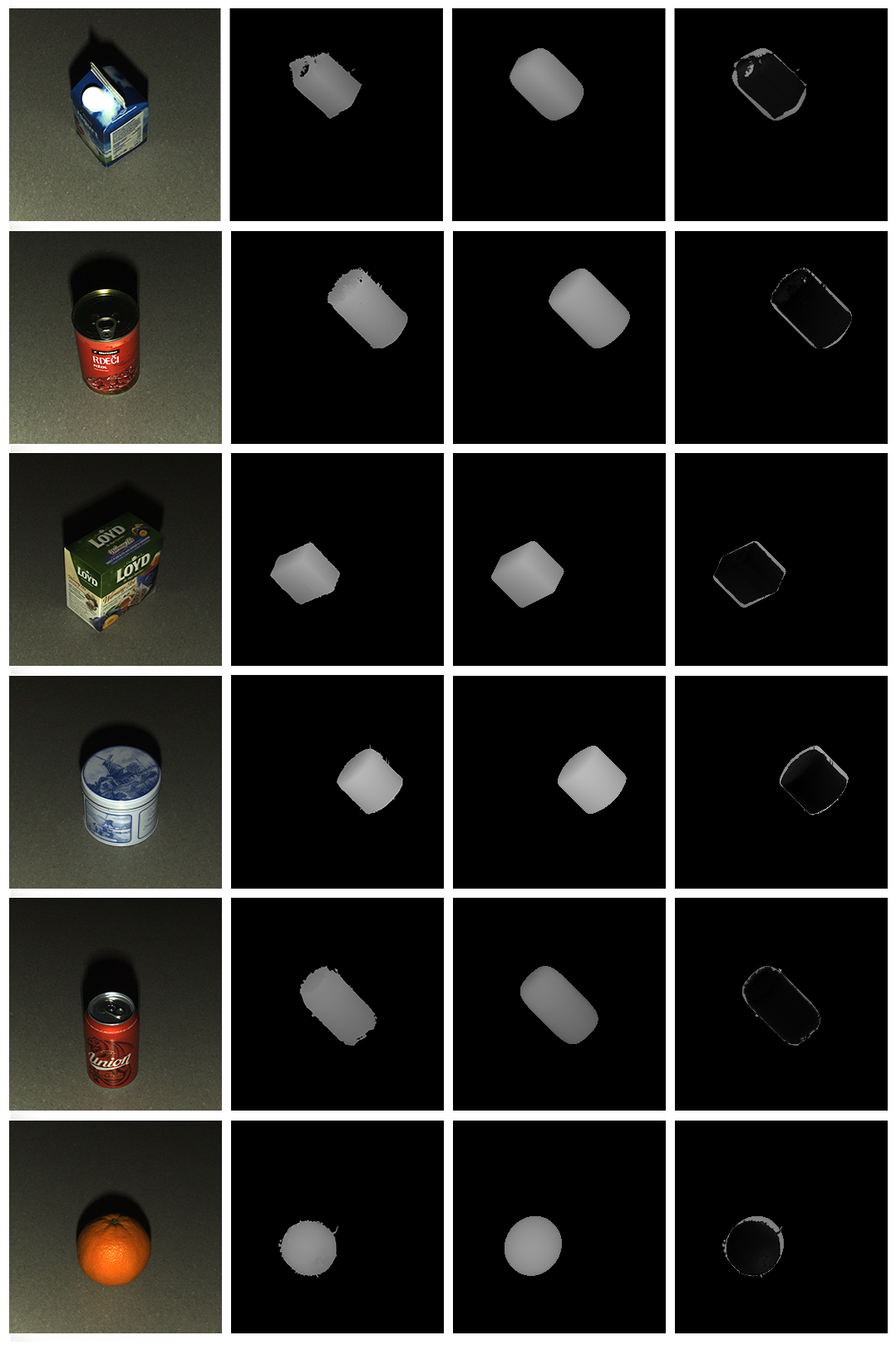}\\\vspace{1mm}
        \caption{Visual examples of the fitting performance of the proposed CNN regressor with real-world data, captured by the Artec MHT 3D scanner. Results are shown for all $6$ scanned objects (in rows). The column show (from left to right): the scanned object in RGB, the rendered range image in an isometric projection, the estimated superquadric, and the absolute difference between the range image and superquadric representation.}
        \label{fig:real_examples}
\end{figure}

We use Artec Studio to manipulate the point cloud data and create a mesh that defines a 3D shape which closely approximates the original object. The point clouds are cleaned using an outlier filter to remove the noise generated by the 3D scanner. We also manually remove all background points during preprocessing, so only the object remains in the processed data. 
Once the preprocessing is completed, we transform the meshes into range images using a mesh manipulation program, called Meshlab. The objects are again rendered in an isometric projection with fixed rotations and the range images are scaled to a size of $256\times256$. We scan a total of $6$ distinct objects 
and show the RGB scans of the objects on the left side of Figure~\ref{fig:real_examples}.  

The generated range images are fed to the CNN regressor, which then produces estimates of the $8$ superquadric parameters for each of the input images. We evaluate the results of this experiment in a qualitative manner and show examples of the recovered 3D shapes in Figure~\ref{fig:real_examples}. Here, the first image in each row shows the scanned object in the form of an RGB image, the second image shows the corresponding range image that represents the input to the CNN regressor, the third image shows the recovered superquadric shape and the last image shows the absolute difference between the input range image and its superquadric representation. We observe that the recovery of superquadric parameters is relatively successful. All of the objects are mostly covered by the generated superquadric, while for most cases the estimated superquadric completely encloses the scanned objects. Even though the CNN regressor was trained on clean data without any additional augmentations, it works considerably well on images where some noise is introduced around object's contour during the scanning process. The model seems to have no issues with objects, where range data is sparse due to reflections. As said, the general shape is fitted successfully, but we can also observe some more complex interactions. For example, both objects in second and fifth row could be approximated trivially using a cylinder shape. Our regressor, however, detects the slight difference in edge roundness and adjusts the shape parameters accordingly. 






\comment{
Quaternion results: 
dims, shape, pos, M, quat
9.42, 0.04, 2.75. 0.67, 0.59
+11.8788  +11.8472   +4.5345   +0.0332   +0.0389   +1.8369   +1.8496   +4.5579   +0.6724   +0.6628  +0.6668   +0.6692   +0.6654   +0.6649   +0.6682   +0.6709   +0.6639   +0.5957   +0.5920   +0.5985   +0.5951

Rotation matrix results: 
dims, shape, pos, M, quat
11.76, 0.24, 7.95, 0.66, 0.45
+11.8128  +11.8448  +11.6280   +0.2396   +0.2488   +4.4241   +4.1823  +15.2458   +0.6683   +0.6683   +0.6632   +0.6653   +0.6594   +0.6621   +0.6636   +0.6670   +0.6741   +0.4463   +0.4476   +0.4479   +0.4467

Isometric projection results: 
dims, shape, pos
1.00, 0.02, 1.13
+1.0147   +1.0244   +0.9649   +0.0150   +0.0180   +0.7033   +0.8592   +1.8338
}

\section{Conclusions\label{sec:conclusion}}

In this preliminary study we introduced a novel CNN-based approach to superquadric recovery from range images. We addressed a constrained recovery problem where a single 3D shape to be modeled by a superquadric was assumed to be present in the data and the superquadric models without rotation were considered. We showed that the proposed CNN model was able to outperform existing state-of-the-art superquadric recovery models in terms of parameter prediction accuracy, but without iterative fitting procedures and at the fraction of the time. To the best of our knowledge, this work was the first to introduce a CNN-based superquadric recovery model and show that this line of research has considerable potential. As part of our future work, we plan to extend our model to more general problems involving rotated superquadrics, where additional ambiguities are introduced into the recovery problem (superquadrics with the same appearance may have different parameters), and complex 3D shapes where a segmentation step is required before the superquadric models can be recovered. 

\section*{Acknowledgements}
This research was supported in parts by the ARRS (Slovenian Research Agency) Project J2-9228 
``A neural network solution to segmentation and recovery of superquadric models from 3D image data'',
ARRS Research Program P2-0250 (B) ``Metrology and Biometric Systems'' and the 
ARRS Research Program P2-0214 (A) ``Computer Vision''. 

\bibliographystyle{IEEEtran}
\bibliography{bibs}

\end{document}